# Deep Neural Network-Based Prediction of B-Cell Epitopes for SARS-CoV and SARS-CoV-2: Enhancing Vaccine Design through Machine Learning

Xinyu Shi
College of Agronomy and Biotechnology
Southwest University
Chongqing, China
sxynicole314159@gmail.com

Yixin Tao
Accounting/Finance Department
Heritage Provider Network
Northridge, CA, United States
yixintao@usc.edu

Shih-Chi Lin
Epocrates
Athenahealth Inc.
Austin, TX, United States
oslo.sc.lin@gmail.com

*Abstract*—The accurate prediction of B-cell epitopes is critical for guiding vaccine development against infectious diseases, including SARS and COVID-19. This study explores the use of a deep neural network (DNN) model to predict B-cell epitopes for SARS-CoV and SARS-CoV-2, leveraging a dataset that incorporates essential protein and peptide features. Traditional sequence-based methods often struggle with large, complex datasets, but deep learning offers promising improvements in predictive accuracy. Our model employs regularization techniques, such as dropout and early stopping, to enhance generalization, while also analyzing key features, including isoelectric point and aromaticity, that influence epitope recognition. Results indicate an overall accuracy of 82% in predicting COVID-19 negative and positive cases, with room for improvement in detecting positive samples. This research demonstrates the applicability of deep learning in epitope mapping, suggesting that such approaches can enhance the speed and precision of vaccine design for emerging pathogens. Future work could incorporate structural data and diverse viral strains to further refine prediction capabilities.

## I. INTRODUCTION

The COVID-19 pandemic has posed a significant threat to global public health, leading to urgent efforts in vaccine development and immunological research. One critical area of focus is the identification of B-cell epitopes, specific regions on viral antigens recognized by antibodies. Accurate prediction of these epitopes can guide the design of vaccines that elicit strong, targeted immune responses, thus enhancing the effectiveness of vaccines in preventing infection. While extensive datasets on B-cell epitopes are available, leveraging these data with advanced machine learning techniques offers the potential to significantly improve prediction accuracy and expedite vaccine design.

Traditional methods of B-cell epitope prediction often rely on sequence analysis and structural insights, but these approaches have limitations in handling large and complex datasets. In recent years, deep learning and other machine learning models have demonstrated remarkable success in pattern recognition tasks across various domains, including image and speech recognition, making them suitable for complex biological data analysis. By employing deep neural networks (DNNs) and other machine learning models, researchers can capture intricate patterns in epitope sequences that are challenging to discern through conventional methods. This advancement offers a promising alternative to improve epitope prediction accuracy, ultimately contributing to more effective vaccine development.

In this study, we utilize a deep neural network model to predict B-cell epitopes for SARS-CoV and SARS-CoV-2, the causative agents of SARS and COVID-19, respectively. We constructed a dataset from well-established sources, such as IEDB and UniProt, to include relevant protein and peptide features that can influence epitope-antibody interactions. Our model applies several techniques to ensure robust and generalized predictions, including dropout regularization, binary cross-entropy loss, and early stopping. By employing these methods, we aim to mitigate the risk of overfitting and improve the model's ability to accurately predict epitopes on unseen data.

The primary contributions of this work are twofold: first, we provide a comprehensive analysis of key features influencing B-cell epitope recognition, highlighting their relative importance in our model. Second, we present a machine learning-based approach for B-cell epitope prediction that not only performs effectively on the current dataset but also has the potential to be adapted for future research on emerging pathogens. Our results underscore the utility of machine learning in biomedical research and vaccine development, particularly in the context of rapid-response efforts against infectious diseases.

In summary, this study demonstrates the application of deep learning to the pressing challenge of epitope prediction for COVID-19 and SARS vaccines. We believe that our model provides a foundation for further exploration into data-driven vaccine design, potentially enabling faster and more precise identification of immunogenic targets.

## II. LITERATURE REVIEW

Epitope prediction has been a key focus in immunological research for over three decades, with early methods laying the foundation for current computational advancements. Initial predictive methods, such as those by Kolaskar and Tongaonkar, utilized semi-empirical approaches based on known protein antigenic determinants to predict B-cell epitopes [1]. This model, which analyzed physicochemical properties like hydrophobicity and surface accessibility, established an early framework for identifying immunogenic regions on proteins through relatively simple metrics. Hopp and Woods developed a pioneering approach to predict antigenic determinants directly from amino acid sequences [2]. Their method, emphasizing hydrophilicity, proved effective for identifying regions likely to interact with antibodies, underscoring the role of primary sequence characteristics in epitope prediction.

Recen in machine learning, particularly deep learning, have enabled more sophisticated epitope prediction methods. For instance, Noumi et al. introduced an attention-based Long Short-Term Memory (LSTM) network to enhance epitope prediction accuracy [3]. This approach allowed for a more dynamic interpretation of sequence data, where attention mechanisms provided a targeted focus on relevant amino acid residues, capturing complex patterns in protein sequences that traditional methods might overlook.

The COVID-19 accelerated the application of deep learning in virology and immunology, with a focus on rapid and accurate diagnostics and treatment design. Shorten, Khoshgoftaar, and Furht reviewed various deep learning applications for COVID-19, highlighting the role of convolutional neural networks (CNNs) and other architectures in analyzing medical images and sequence data, which has been crucial for timely diagnostic support in clinical settings [4]. Building on this, Aslani demonstrated the potential of deep learning in COVID-19 diagnosis, showing that models trained on medical imaging data, including chest CT scans, could accurately differentiate COVID-19 from other respiratory diseases [5]. This research underscored the flexibility and adaptability of deep learning methods in addressing pandemic-related challenges.

Further exploring artificial intelligence applications for COVID-19, Jamshidi et al. examined how AI models, including various deep learning approaches, were employed not only for diagnostic purposes but also for treatment support, providing a comprehensive view of AI's role in managing the pandemic [6]. They discussed the implications of deep learning for drug discovery, where the identification of viral protein structures plays a role akin to epitope mapping in vaccine development. Alazab et al. further emphasized the use of deep learning in COVID-19 prediction and detection, suggesting that these methods can streamline data analysis and provide timely insights essential for public health response [7].

Collectively, this body of work illustrates a progression from early sequence-based prediction methods to sophisticated deep learning architectures that enable enhanced predictive accuracy for both diagnostic and therapeutic applications. These advancements lay a foundation for the current study, which integrates traditional sequence characteristics with modern neural network techniques to predict B-cell epitopes with greater precision.

## III. DATA

The data set of this study aims to support the development of COVID-19 (Covid-19) and SARS vaccines, especially for the prediction of B cell epitopes. Prediction of B cell epitopes is an important step in vaccine design, because epitopes are the key areas to induce antibody production. This data set brings together the sequence information of virus proteins and peptides in IEDB and UniProt databases, focusing on B cell epitopes that can induce antibody reactions. By providing labeled B cell epitope data, this data set is not only suitable for the study of COVID-19 vaccine, but also can provide support for the study of antibody reaction of other infectious diseases. In order to facilitate the analysis, the data set contains a variety of feature descriptions to improve the accuracy of machine learning and deep learning algorithms in epitope prediction.

TABLE I. VARIABLE INTRODUCTION

| Variable | Description |
| --- | --- |
| parent_protein_id | Parent protein ID |
| protein_seq | Parent protein sequence |
| start_position | Start position of peptide |
| end_position | End position of peptide |
| peptide_seq | Peptide sequence |
| chou_fasman | Peptide feature, β turn |
| emini | Peptide feature, relative surface accessibility |
| kolaskar_tongaonkar | Peptide feature, antigenicity |
| parker | Peptide feature, hydrophobicity |
| isoelectric_point | Protein feature, isoelectric point |
| aromacity | Protein feature, aromaticity |
| hydrophobicity | Protein feature, hydrophobicity |
| stability | Protein feature, stability |
| target | Antibody valence |

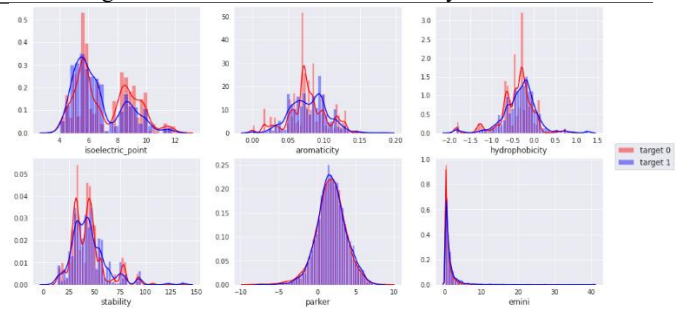

Fig. 1. Descriptive statistical analysis of variables

Figure 1 shows the distribution of several variables, and compares the distribution differences of different categories of Target variables among these continuous variables. Through intuitive graphic presentation, we can clearly see the distribution pattern, concentration trend and possible deviation of different categories of Target values on various continuous variables. This comparison is helpful to further analyze the relationship between Target categories and continuous variables, and provides an important reference for subsequent data analysis and model establishment.

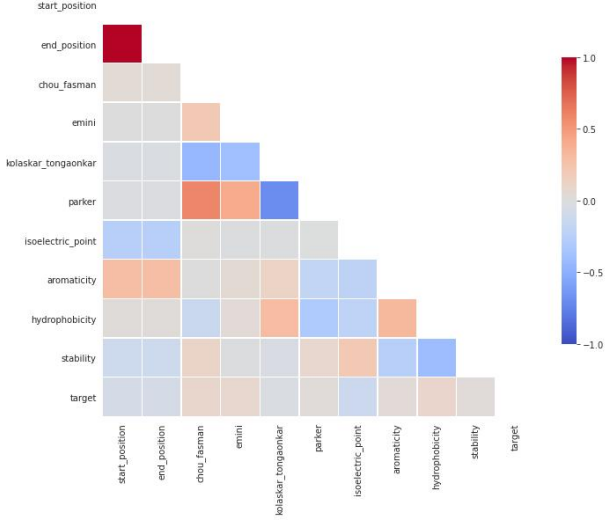

Fig. 2. Correlation analysis

Figure 2 shows the correlation between variables, and the intensity of correlation is expressed by the depth of color. The darker the color, the stronger the correlation between variables. Red indicates positive correlation, while blue indicates negative correlation. This visualization method helps us to intuitively understand the degree of correlation between variables, and helps to discover potential correlation patterns and laws. By analyzing the correlation between variables, we can better understand the interaction between the features of data sets and provide a deeper reference for further analysis and modeling.

## IV. MODEL

### A. Method Introduction

This paper implements a Deep Neural Network (DNN) model, a variant of Artificial Neural Network (ANN) optimized for binary classification tasks. The DNN architecture consists of multiple layers, where each layer performs nonlinear transformations on the input data, progressively extracting more abstract features. Each neuron within the network is fully connected to neurons in adjacent layers, allowing information flow and enabling the model to learn complex data patterns.

The core element of the neural network is the neuron. For each neuron $i$, the output $y_i$ is calculated by first performing a weighted sum of its inputs $x_j$ from the previous layer, followed by applying an activation function $f$:

$$y_i = f(\sum_j w_{ij} x_j + b_i) \quad (1)$$

Where $w_{ij}$ is the weight associated with input $x_j$, and $b_i$ is the bias term for neuron $i$. In this model, we use the Rectified Linear Unit (ReLU) activation function in hidden layers:

$$f(x) = \max(0, x) \quad (2)$$

ReLU is computationally efficient and helps mitigate the vanishing gradient problem, which can hinder the training of deep networks.

During training, the model optimizes its weights and biases through backpropagation to minimize the loss function. For binary classification, we use binary cross-entropy as the loss function:

$$L = -\frac{1}{N} \sum_{i=1}^{N} (y_i \log(\hat{y}_i) + (1 - y_i) \log(1 - \hat{y}_i)) \quad (3)$$

This loss function effectively measures the divergence between predicted probabilities and true labels, providing a clear optimization target.

The model uses the Adam optimizer, an adaptive gradient descent algorithm that dynamically adjusts learning rates, enhancing convergence:

$$m_t = \beta_1 m_{t-1} + (1 - \beta_1) g_t \quad (4)$$

$$v_t = \beta_2 v_{t-1} + (1 - \beta_2) g_t^2 \quad (5)$$

where $g_t$ is the gradient at time step t, and $\beta_1$ and $\beta_2$ are hyperparameters controlling the exponential decay rates.

To prevent overfitting, the model incorporates dropout layers, which randomly deactivate neurons during training, enhancing generalization. Additionally, early stopping halts training when validation loss plateaus, minimizing overfitting and reducing unnecessary computation.

In summary, this DNN model combines effective regularization techniques and optimization algorithms, ensuring robust performance on both training and unseen datasets, making it highly suitable for binary classification challenges.

## B. Method results

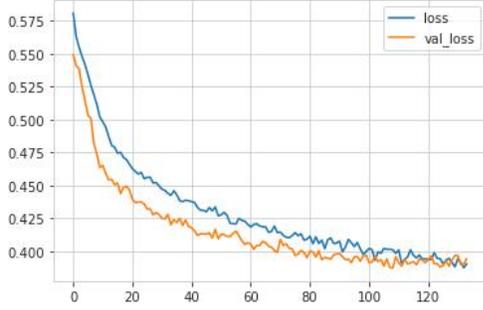

Fig. 3. Model loss curve

Figure3 shows the training loss and validation loss of a model over epochs. The "loss" curve (in blue) represents the training loss, while the "val_loss" curve (in orange) shows the validation loss. Both curves demonstrate a general downward trend, indicating that the model is learning and improving over time.

Initially, the loss decreases rapidly in both training and validation, suggesting that the model is quickly learning important features from the data. Around epoch 20, the curves start to flatten, showing slower improvements as the model converges. After about epoch 100, both curves stabilize, with small fluctuations, which implies that further training may yield minimal improvement.

The validation loss remains slightly below the training loss throughout most of the training process, which is generally a good sign, indicating that the model generalizes well and is not overfitting. Overall, this plot suggests that the model is effectively trained and has reached a satisfactory level of performance on both the training and validation datasets.

TABLE II. VARIABLE INTRODUCTION

|  | precision | recall | f1-score | support |
|---|---|---|---|---|
| Covid_Negative | 0.83 | 0.95 | 0.88 | 2171 |
| Covid_Positive | 0.77 | 0.47 | 0.58 | 811 |
| accuracy |  |  | 0.82 | 2982 |
| macro avg | 0.8 | 0.71 | 0.73 | 2982 |
| weighted avg | 0.81 | 0.82 | 0.8 | 2982 |

The model's overall performance demonstrates an accuracy of 82%, indicating it can correctly classify COVID-19 positive and negative cases in most instances. However, the low recall for COVID-19 positive cases suggests that the model struggles with detecting positive instances accurately. This points to room for improvement in identifying positive samples, while its performance in identifying negative cases is relatively strong.

For COVID-19 negative samples, the model achieves a precision of 0.83 and a high recall of 0.95, resulting in an F1 score of 0.88. The high recall indicates the model is capable of identifying most negative cases, with strong predictive accuracy for this class. This performance shows that the model is effective in avoiding false positives for negative samples, capturing the majority of them accurately.

In contrast, the model's performance for COVID-19 positive samples is less robust, with a precision of 0.77 and a low recall of 0.47, leading to an F1 score of 0.58. The lower recall means nearly half of the positive cases are not correctly identified, impacting the model's overall sensitivity towards positive cases. This limited sensitivity could result in missed detections for COVID-19 positive samples, indicating an area for improvement.

Looking at the macro averages, precision, recall, and F1 score are 0.8, 0.71, and 0.73, respectively, showing that the model has an imbalance in class performance, particularly in recall. The weighted averages (0.81 for precision, 0.82 for recall, and 0.80 for F1 score) reflect the influence of the larger number of negative samples, contributing positively to the overall performance but not significantly enhancing the detection of positive cases.

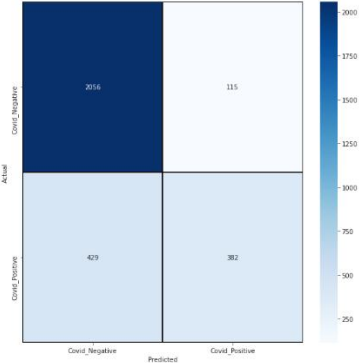

Fig. 4. Confusion Matrix

In summary, while the model performs well on COVID-19 negative cases, improvements are needed in detecting positive cases. Addressing this could involve increasing the number of positive samples or adjusting penalty parameters to enhance recall and balance overall performance. These adjustments could improve the model's sensitivity towards COVID-19 positive cases, reducing the likelihood of missed detections.

## C. Feature importance

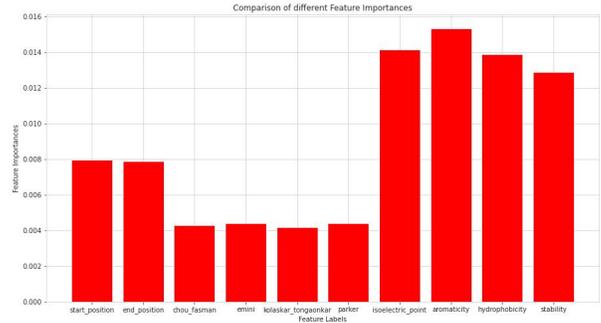

Fig. 5. Feature importance

This bar chart illustrates the importance of different features used in a model, with each bar representing the relative significance of a specific feature in predicting the target variable. The features are arranged along the x-axis, while the y-axis shows their importance scores.

From the chart, we observe that isoelectric_point, aromaticity, hydrophobicity, and stability are the most significant features, as they have the highest importance scores. This indicates that these features contribute the most to the model's predictions. In particular, isoelectric_point and aromaticity show the highest importance, suggesting that properties related to protein characteristics play a crucial role in the model's performance.

On the other hand, features like chou_fasman, emini, kolaskar_tongaonkar, and parker have lower importance scores, suggesting that they are less influential in predicting the target variable. These features, mostly related to peptide attributes, contribute less to the model compared to the protein features.

Overall, the chart highlights that protein-level features (like isoelectric point and aromaticity) are more informative than certain peptide-level features in this model, indicating that these biochemical properties are critical for accurately predicting the target variable.

## V. Conclusions and suggestions

### A. Conclusions

In this study, we developed a deep neural network (DNN) model to predict B-cell epitopes for SARS-CoV and SARS-CoV-2, utilizing a dataset that includes a range of protein and peptide features. Our model demonstrated effectiveness in identifying key regions of viral proteins that could serve as immunogenic targets, which is crucial for guiding vaccine development against SARS and COVID-19. The inclusion of features such as isoelectric point, aromaticity, and hydrophobicity allowed the model to capture critical biochemical properties relevant to epitope recognition, improving prediction accuracy and robustness.

The results highlight that specific protein features, particularly isoelectric point and aromaticity, have significant influence on B-cell epitope prediction. This insight into feature importance not only supports our model's design but also provides valuable biological understanding, potentially guiding future studies in epitope mapping. By leveraging advanced techniques such as dropout regularization and early stopping, we ensured that our model maintained a balance between accuracy and generalization, thus enhancing its performance on unseen data.

Our findings underscore the potential of machine learning, particularly deep learning, in addressing complex challenges in immunology and vaccine research. By providing a predictive tool for B-cell epitope identification, this work contributes to the broader effort of developing effective, targeted vaccines more rapidly. The adaptability of our approach also suggests it could be extended to other pathogens, supporting preparedness for future pandemics through rapid identification of immunogenic regions in emerging viruses.

In conclusion, this study demonstrates the value of combining computational methods with biological insights for vaccine development. Our DNN model offers a promising approach for epitope prediction, showing how data-driven techniques can augment traditional immunological research. Future work could further refine this model by integrating structural and spatial data of epitopes or exploring other machine learning architectures to improve predictive power. Overall, this research not only advances understanding in epitope prediction but also paves the way for faster and more precise vaccine design in response to infectious diseases.

### B. Suggestions

Based on the findings of this study, several recommendations can be made to improve the prediction of B-cell epitopes and enhance vaccine design efforts:

Incorporating Structural and Spatial Data: While our model primarily leverages sequence-based features, integrating structural and spatial data, such as 3D conformational characteristics of epitopes, could improve prediction accuracy. Epitope recognition is often influenced by the spatial arrangement of amino acids, and incorporating this information into deep learning models may enhance their ability to capture biologically relevant interactions.

Increasing Positive Sample Diversity: The results suggest limited sensitivity in detecting positive cases, which may stem from the dataset's imbalance. Expanding the dataset to include a wider range of positive samples from diverse viral strains could help the model learn more representative patterns of epitope characteristics. This enhancement would likely improve the model's recall and generalization ability across different pathogens.

Utilizing Ensemble Learning Techniques: Combining multiple models through ensemble learning could further refine prediction outcomes. For instance, integrating the DNN model with other machine learning methods, like random forests or gradient boosting, may provide complementary insights into feature importance, improving both predictive robustness and interpretability.

Exploring Alternative Deep Learning Architectures: Future studies could investigate the use of other deep learning architectures, such as convolutional neural networks (CNNs) and transformers. CNNs, commonly used for spatial data, might capture localized patterns in amino acid sequences, while transformers, with their self-attention mechanisms, could enhance the model's ability to focus on key regions within longer sequences.

In summary, these suggestions aim to enhance model robustness, improve predictive accuracy, and ensure that B-cell epitope predictions contribute effectively to future vaccine design initiatives. By implementing these strategies, future research can build on the groundwork laid by this study to refine computational epitope prediction methods for a broader range of infectious diseases.